# Markov Logic in Infinite Domains

**Parag Singla    Pedro Domingos**
Department of Computer Science and Engineering
University of Washington
Seattle, WA 98195-2350, U.S.A.
{*parag, pedrod*}@cs.washington.edu

## Abstract

Combining first-order logic and probability has long been a goal of AI. Markov logic (Richardson & Domingos, 2006) accomplishes this by attaching weights to first-order formulas and viewing them as templates for features of Markov networks. Unfortunately, it does not have the full power of first-order logic, because it is only defined for finite domains. This paper extends Markov logic to infinite domains, by casting it in the framework of Gibbs measures (Georgii, 1988). We show that a Markov logic network (MLN) admits a Gibbs measure as long as each ground atom has a finite number of neighbors. Many interesting cases fall in this category. We also show that an MLN admits a unique measure if the weights of its non-unit clauses are small enough. We then examine the structure of the set of consistent measures in the non-unique case. Many important phenomena, including systems with phase transitions, are represented by MLNs with non-unique measures. We relate the problem of satisfiability in first-order logic to the properties of MLN measures, and discuss how Markov logic relates to previous infinite models.

## 1 Introduction

Most AI problems are characterized by both uncertainty and complex structure, in the form of multiple interacting objects and relations. Handling both requires combining the capabilities of probabilistic models and first-order logic. Attempts to achieve this have a long history, and have gathered steam in recent years. Within AI, Nilsson (1986) is an early example. Bacchus (1990), Halpern (1990) and coworkers (e.g., Bacchus *et al.* (1996)) produced a substantial body of relevant theoretical work. Around the same time, several authors began using logic programs to compactly specify complex Bayesian networks, an approach known as knowledge-based model construction (Wellman et al., 1992). More recently, many combinations of (subsets of) first-order logic and probability have been proposed in the burgeoning field of statistical relational learning (Getoor & Taskar, 2007), including probabilistic relational models (Friedman et al., 1999), stochastic logic programs (Muggleton, 1996), Bayesian logic programs (Kersting & De Raedt, 2001), and others.

One of the most powerful representations to date is Markov logic (Richardson & Domingos, 2006). Markov logic is a simple combination of Markov networks and first-order logic: each first-order formula has an associated weight, and each grounding of a formula becomes a feature in a Markov network, with the corresponding weight. The use of Markov networks instead of Bayesian networks obviates the difficult problem of avoiding cycles in all possible groundings of a relational model (Taskar et al., 2002). The use of first-order logic instead of more limited representations (e.g., description logics, Horn clauses) makes it possible to compactly represent a broader range of dependencies. For example, a dependency between relations like "Friends of friends are (usually) friends" cannot be specified compactly in (say) probabilistic relational models, but in Markov logic it suffices to write down the corresponding formula and weight. Markov logic has been successfully applied in a variety of domains (Domingos et al., 2006), and open source software with implementations of state-of-the-art inference and learning algorithms for it is available (Kok et al., 2006).

One limitation of Markov logic is that it is only defined for finite domains. While this is seldom a problem in practice, considering the infinite limit can simplify the treatment of some problems, and yield new insights. We would also like to elucidate how far it is possible to combine the full power of first-order logic and graphical models. Thus in this paper we extend Markov logic to infinite domains. Our treatment is based on the theory of Gibbs measures (Georgii, 1988). Gibbs measures are infinite-dimensional extensions of Markov networks, and have been studied extensively by statistical physicists and mathematical statisti-cians, due to their importance in modeling systems with phase transitions. We begin with some necessary background on first-order logic and Gibbs measures. We then define MLNs over infinite domains, state sufficient conditions for the existence and uniqueness of a probability measure consistent with a given MLN, and examine the important case of MLNs with non-unique measures. Next, we establish a correspondence between the problem of satisfiability in logic and the existence of MLN measures with certain properties. We conclude with a discussion of the relationship between infinite MLNs and previous infinite relational models.

## 2 Background

### 2.1 First-Order Logic

A *first-order knowledge base* is a set of sentences or formulas in first-order logic (Genesereth & Nilsson, 1987). Formulas are constructed using four types of symbols: constants, variables, functions, and predicates. Constant symbols represent objects in the domain of discourse (e.g., people: Anna, Bob, Chris, etc.). Variable symbols range over the objects in the domain (or a subset of it, in which case they are *typed*). Function symbols (e.g., MotherOf) represent mappings from tuples of objects to objects. Predicate symbols represent relations among objects (e.g., Friends) or attributes of objects (e.g., Smokes). A *term* is any expression representing an object. It can be a constant, a variable, or a function applied to a tuple of terms. For example, Anna, x, and GreatestCommonDivisor(x, y) are terms. An *atomic formula* or *atom* is a predicate symbol applied to a tuple of terms (e.g., Friends(x, MotherOf(Anna))). A *ground term* is a term containing no variables. A *ground atom* or *ground predicate* is an atomic formula all of whose arguments are ground terms. Formulas are recursively constructed from atomic formulas using logical connectives and quantifiers. A *positive literal* is an atomic formula; a *negative literal* is a negated atomic formula. A *clause* is a disjunction of literals. Every first-order formula can be converted into an equivalent formula in *prenex conjunctive normal form*, $Qx_1 \ldots Qx_n C(x_1, \ldots, x_n)$, where each $Q$ is a quantifier, the $x_i$ are the quantified variables, and $C(\ldots)$ is a conjunction of clauses.

The *Herbrand universe* $\mathbf{U(C)}$ of a set of clauses $\mathbf{C}$ is the set of all ground terms constructible from the function and constant symbols in $\mathbf{C}$ (or, if $\mathbf{C}$ contains no constants, some arbitrary constant, e.g., A). If $\mathbf{C}$ contains function symbols, $\mathbf{U(C)}$ is infinite. (For example, if $\mathbf{C}$ contains solely the function f and no constants, $\mathbf{U(C)}$ = {f(A), f(f(A)), f(f(f(A))), ...}.) Some authors define the *Herbrand base* $\mathbf{B(C)}$ of $\mathbf{C}$ as the set of all ground atoms constructible from the predicate symbols in $\mathbf{C}$ and the terms in $\mathbf{U(C)}$. Others define it as the set of all ground clauses constructible from the clauses in $\mathbf{C}$ and the terms in $\mathbf{U(C)}$. For convenience, in this paper we will define it as the union of the two, and talk about the *atoms in* $\mathbf{B(C)}$ and *clauses in* $\mathbf{B(C)}$ as needed.

An *interpretation* is a mapping between the constant, predicate and function symbols in the language and the objects, functions and relations in the domain. In a *Herbrand interpretation* there is a one-to-one mapping between ground terms and objects (i.e., every object is represented by some ground term, and no two ground terms correspond to the same object). A *model* or *possible world* specifies which relations hold true in the domain. Together with an interpretation, it assigns a truth value to every atomic formula, and thus to every formula in the knowledge base.

### 2.2 Gibbs Measures

Gibbs measures are infinite-dimensional generalizations of Gibbs distributions. A Gibbs distribution, also known as a log-linear model or exponential model, and equivalent under mild conditions to a Markov network or Markov random field, assigns to a state $\mathbf{x}$ the probability

$$P(\mathbf{X} = \mathbf{x}) = \frac{1}{Z} \exp\left(\sum_i w_i f_i(\mathbf{x})\right) \qquad (1)$$

where $w_i$ is any real number, $f_i$ is an arbitrary function or *feature* of $\mathbf{x}$, and $Z$ is a normalization constant. In this paper we will be concerned exclusively with Boolean states and functions (i.e., states are binary vectors, corresponding to possible worlds, and functions are logical formulas). Markov logic can be viewed as the use of first-order logic to compactly specify families of these functions (Richardson & Domingos, 2006). Thus, a natural way to generalize it to infinite domains is to use the existing theory of Gibbs measures (Georgii, 1988). Although Gibbs measures were primarily developed to model regular lattices (e.g., ferromagnetic materials, gas/liquid phases, etc.), the theory is quite general, and applies equally well to the richer structures definable using Markov logic.

One problem with defining probability distributions over infinite domains is that the probability of most or all worlds will be zero. Measure theory allows us to overcome this problem by instead assigning probabilities to sets of worlds (Billingsley, 1995). Let $\Omega$ denote the set of all possible worlds, and $\mathcal{E}$ denote a set of subsets of $\Omega$. $\mathcal{E}$ must be a $\sigma$-algebra, i.e., it must be non-empty and closed under complements and countable unions. A function $\mu : \mathcal{E} \to \mathbb{R}$ is said to be a *probability measure* over $(\Omega, \mathcal{E})$ if $\mu(E) \geq 0$ for every $E \in \mathcal{E}$, $\mu(\Omega) = 1$, and $\mu(\bigcup E_i) = \sum \mu(E_i)$, where the union is taken over any countable collection of disjoint elements of $\mathcal{E}$.

A related difficulty is that in infinite domains the sum in Equation 1 may not exist. However, the distribution of any finite subset of the state variables conditioned on its complement is still well defined. We can thus define the infinite distribution indirectly by means of an infinite collection of finite conditional distributions. This is the basic idea in Gibbs measures.

Let us introduce some notation which will be used throughout the paper. Consider a countable set of variables $\mathbf{S} = \{X_1, X_2, \ldots\}$, where each $X_i$ takes values in $\{0, 1\}$. Let $\mathbf{X}$ be a finite set of variables in $\mathbf{S}$, and $\mathbf{S_X} = \mathbf{S} \setminus \mathbf{X}$. A possible world $\omega \in \Omega$ is an assignment to all the variables in $\mathbf{S}$. Let $\omega_{\mathbf{X}}$ denote the assignment to the variables in $\mathbf{X}$ under $\omega$, and $\omega_{X_i}$ the assignment to $X_i$. Let $\mathcal{X}$ denote the set of all finite subsets of $\mathbf{S}$. A *basic event* $\mathbf{X} = \mathbf{x}$ is an assignment of values to a finite subset of variables $\mathbf{X} \in \mathcal{X}$, and denotes the set of possible worlds $\omega \in \Omega$ such that $w_{\mathbf{X}} = \mathbf{x}$. Let $\mathbf{E}$ be the set of all basic events, and let $\mathcal{E}$ be the $\sigma$-algebra generated by $\mathbf{E}$, i.e., the smallest $\sigma$-algebra containing $\mathbf{E}$. An element $E$ of $\mathcal{E}$ is called an *event*, and $\mathcal{E}$ is the *event space*. The following treatment is adapted from Georgii (1988).

**Definition 1.** *An* interaction potential *(or simply a* potential*) is a family* $\Phi = (\Phi_{\mathbf{V}})_{\mathbf{V} \in \mathcal{X}}$ *of functions* $\Phi_{\mathbf{V}} : \mathbf{V} \to \mathbb{R}$ *such that, for all* $\mathbf{X} \in \mathcal{X}$ *and* $\omega \in \Omega$*, the summation*

$$H_{\mathbf{X}}^{\Phi}(\omega) = \sum_{\mathbf{V} \in \mathcal{X}, \mathbf{V} \cap \mathbf{X} \neq \emptyset} \Phi_{\mathbf{V}}(\omega_{\mathbf{V}}) \quad (2)$$

*is finite.* $H_{\mathbf{X}}^{\Phi}$ *is called the Hamiltonian in* $\mathbf{X}$ *for* $\Phi$.

Intuitively, the Hamiltonian $H_{\mathbf{X}}^{\Phi}$ includes a contribution from all the potentials $\Phi_{\mathbf{V}}$ which share at least one variable with the set $\mathbf{X}$. Given an interaction potential $\Phi$ and a subset of variables $\mathbf{X}$, we define the conditional distribution $\gamma_{\mathbf{X}}^{\Phi}(\mathbf{X}|\mathbf{S_X})$ as[1]

$$\gamma_{\mathbf{X}}^{\Phi}(\mathbf{X} = \mathbf{x}|\mathbf{S_X} = \mathbf{y}) = \frac{\exp(H_{\mathbf{X}}^{\Phi}(\mathbf{x}, \mathbf{y}))}{\sum_{\mathbf{x} \in \mathrm{Dom}(\mathbf{X})} \exp(H_{\mathbf{X}}^{\Phi}(\mathbf{x}, \mathbf{y}))} \quad (3)$$

where the denominator is called the *partition function* in $\mathbf{X}$ for $\Phi$ and denoted by $Z_{\mathbf{X}}^{\Phi}$, and $\mathrm{Dom}(\mathbf{X})$ is the domain of $\mathbf{X}$. Equation 3 can be easily extended to arbitrary events $E \in \mathcal{E}$ by defining $\gamma_{\mathbf{X}}^{\Phi}(E|\mathbf{S_X})$ to be non-zero only when $E$ is consistent with the assignment in $\mathbf{S_X}$. Details are skipped here to keep the discussion simple, and can be found in Georgii (1988). The family of conditional distributions $\gamma^{\Phi} = (\gamma_{\mathbf{X}}^{\Phi})_{\mathbf{X} \in \mathcal{X}}$ as defined above is called a *Gibbsian specification*.[2]

Given a measure $\mu$ over $(\Omega, \mathcal{E})$ and conditional probabilities $\gamma_{\mathbf{X}}^{\Phi}(E|\mathbf{S_X})$, let the composition $\mu\gamma_{\mathbf{X}}^{\Phi}$ be defined as

$$\mu\gamma_{\mathbf{X}}^{\Phi}(E) = \int_{\mathrm{Dom}(\mathbf{S_X})} \gamma_{\mathbf{X}}^{\Phi}(E|\mathbf{S_X}) \, \partial\mu \quad (4)$$

---

[1]For physical reasons, this equation is usually written with a negative sign in the exponent, i.e., $\exp[-H_{\mathbf{X}}^{\Phi}(\omega)]$. Since this is not relevant in Markov logic and does not affect any of the results, we omit it.

[2]Georgii (1988) defines Gibbsian specifications in terms of underlying independent specifications. For simplicity, we assume these to be equidistributions and omit them throughout this paper.

$\mu\gamma_{\mathbf{X}}^{\Phi}(E)$ is the probability of event $E$ according to the conditional probabilities $\gamma_{\mathbf{X}}^{\Phi}(E|\mathbf{S_X})$ and the measure $\mu$ on $\mathbf{S_X}$. We are now ready to define Gibbs measure.

**Definition 2.** *Let* $\gamma^{\Phi}$ *be a Gibbsian specification. Let* $\mu$ *be a probability measure over the measurable space* $(\Omega, \mathcal{E})$ *such that, for every* $\mathbf{X} \in \mathcal{X}$ *and* $E \in \mathcal{E}$*,* $\mu(E) = \mu\gamma_{\mathbf{X}}^{\Phi}(E)$*. Then the specification* $\gamma^{\Phi}$ *is said to admit the* Gibbs *measure* $\mu$*. Further,* $\mathcal{G}(\gamma^{\Phi})$ *denotes the set of all such measures.*

In other words, a Gibbs measure is consistent with a Gibbsian specification if its event probabilities agree with those obtained from the specification. Given a Gibbsian specification, we can ask whether there exists a Gibbs measure consistent with it ($|\mathcal{G}(\gamma^{\Phi})| > 0$), and whether it is unique ($|\mathcal{G}(\gamma^{\Phi})| = 1$). In the non-unique case, we can ask what the structure of $\mathcal{G}(\gamma^{\Phi})$ is, and what the measures in it represent. We can also ask whether Gibbs measures with specific properties exist. The theory of Gibbs measures addresses these questions. In this paper we apply it to the case of Gibbsian specifications defined by MLNs.

## 3 Infinite MLNs

### 3.1 Definition

A Markov logic network (MLN) is a set of weighted first-order formulas. As we saw in the previous section, these can be converted to equivalent formulas in prenex CNF. We will assume throughout that all existentially quantified variables have finite domains, unless otherwise specified. While this is a significant restriction, it still includes essentially all previous probabilistic relational representations as special cases. Existentially quantified formulas can now be replaced by finite disjunctions. By distributing conjunctions over disjunctions, every prenex CNF can now be converted to a quantifier-free CNF, with all variables implicitly universally quantified.

The Herbrand universe $\mathbf{U}(\mathbf{L})$ of an MLN $\mathbf{L}$ is the set of all ground terms constructible from the constants and function symbols in the MLN. The Herbrand base $\mathbf{B}(\mathbf{L})$ of $\mathbf{L}$ is the set of all ground atoms and clauses constructible from the predicates in $\mathbf{L}$, the clauses in the CNF form of $\mathbf{L}$, and the terms in $\mathbf{U}(\mathbf{L})$, replacing typed variables only by terms of the corresponding type. We assume Herbrand interpretations throughout. We are now ready to formally define MLNs.

**Definition 3.** *A* Markov logic network (MLN) $\mathbf{L}$ *is a (finite) set of pairs* $(F_i, w_i)$*, where* $F_i$ *is a formula in first-order logic and* $w_i$ *is a real number.* $\mathbf{L}$ *defines a countable set of variables* $\mathbf{S}$ *and interaction potential* $\Phi^{\mathbf{L}} = (\Phi_{\mathbf{X}}^{\mathbf{L}})_{\mathbf{X} \in \mathcal{X}}$*,* $\mathcal{X}$ *being the set of all finite subsets of* $\mathbf{S}$*, as follows:*

1. $\mathbf{S}$ *contains a binary variable for each atom in* $\mathbf{B}(\mathbf{L})$*.*

*The value of this variable is 1 if the atom is true, and 0 otherwise.*

2. $\Phi_{\mathbf{X}}^{\mathbf{L}}(\mathbf{x}) = \sum_j w_j f_j(\mathbf{x})$, *where the sum is over the clauses $C_j$ in $\mathbf{B}(\mathbf{L})$ whose arguments are exactly the elements of $\mathbf{X}$. If $F_{i(j)}$ is the formula in $\mathbf{L}$ from which $C_j$ originated, and $F_{i(j)}$ gave rise to $n$ clauses in the CNF form of $\mathbf{L}$, then $w_j = w_i/n$. $f_j(\mathbf{x}) = 1$ if $C_j$ is true in world $\mathbf{x}$, and $f_j = 0$ otherwise.*

For $\Phi^{\mathbf{L}}$ to correspond to a well-defined Gibbsian specification, the corresponding Hamiltonians (Equation 2) need to be finite. This brings us to the following definition.

**Definition 4.** *Let $\mathbf{C}$ be a set of first-order clauses. Given a ground atom $X \in \mathbf{B}(\mathbf{C})$, let the neighbors $\mathbf{N}(X)$ of $X$ be the atoms that appear with it in some ground clause. $\mathbf{C}$ is said to be* locally finite *if each atom in the Herbrand base of $\mathbf{C}$ has a finite number of neighbors, i.e., $\forall X \in \mathbf{B}(\mathbf{C})$, $|\mathbf{N}(X)| < \infty$. An MLN (or knowledge base) is said to be* locally finite *if the set of its clauses is locally finite.*

It is easy to see that local finiteness is sufficient to ensure a well-defined Gibbsian specification. Given such an MLN $\mathbf{L}$, the distribution $\gamma_{\mathbf{X}}^{\mathbf{L}}$ of a set of variables $\mathbf{X} \in \mathcal{X}$ conditioned on its complement $\mathbf{S}_{\mathbf{X}}$ is given by

$$\gamma_{\mathbf{X}}^{\mathbf{L}}(\mathbf{X}=\mathbf{x}|\mathbf{S}_{\mathbf{X}}=\mathbf{y}) = \frac{\exp\left(\sum_j w_j f_j(\mathbf{x},\mathbf{y})\right)}{\sum_{\mathbf{x}'\in\mathrm{Dom}(\mathbf{X})} \exp\left(\sum_j w_j f_j(\mathbf{x}',\mathbf{y})\right)} \quad (5)$$

where the sum is over the clauses in $\mathbf{B}(\mathbf{L})$ that contain at least one element of $\mathbf{X}$, and $f_j(\mathbf{x},\mathbf{y}) = 1$ if clause $C_j$ is true under the assignment $(\mathbf{x},\mathbf{y})$ and 0 otherwise. The corresponding Gibbsian specification is denoted by $\gamma^{\mathbf{L}}$.

For an MLN to be locally finite, it suffices that it be $\sigma$-determinate.

**Definition 5.** *A clause is* $\sigma$-determinate *if all the variables with infinite domains it contains appear in all literals.[3] A set of clauses is* $\sigma$-determinate *if each clause in the set is $\sigma$-determinate. An MLN is* $\sigma$-determinate *if the set of its clauses is $\sigma$-determinate.*

Notice that this definition does not require that all literals have the same infinite arguments; for example, the clause $\mathbb{Q}(\mathbf{x}, \mathbf{y}) \Rightarrow \mathbb{R}(\mathbf{f}(\mathbf{x}), \mathbf{g}(\mathbf{x}, \mathbf{y}))$ is $\sigma$-determinate. In essence, $\sigma$-determinacy requires that the neighbors of an atom be defined by functions of its arguments. Because functions can be composed indefinitely, the network can be infinite; because first-order clauses have finite length, $\sigma$-determinacy ensures that neighborhoods are still finite.

---

[3]This is related to the notion of a *determinate clause* in logic programming. In a determinate clause, the grounding of the variables in the head determines the grounding of all the variables in the body. In infinite MLNs, any literal in a clause can be inferred from the others, not just the head from the body, so we require that the (infinite-domain) variables in each literal determine the variables in the others.

If the MLN contains no function symbols, Definition 3 reduces to the one in Richardson and Domingos (2006), with $C$ being the constants appearing in the MLN. This can be easily seen by substituting $\mathbf{X} = \mathbf{S}$ in Equation 5. Notice it would be equally possible to define features for conjunctions of clauses, and this may be preferable for some applications.

### 3.2 Existence

Let $\mathbf{L}$ be a locally finite MLN. The focus of this section is to show that its specification $\gamma^{\mathbf{L}}$ always admits some measure $\mu$. It is useful to first gain some intuition as to why this might not always be the case. Consider an MLN stating that each person is loved by exactly one person: $\forall x \; \exists! y \; \mathtt{Loves}(y, x)$. Let $\omega_k$ denote the event $\mathtt{Loves}(\mathtt{P}_k, \mathtt{Anna})$, i.e., Anna is loved by the $k$th person in the (countably infinite) domain. Then, in the limit of infinite weights, one would expect that $\mu(\bigcup_k \omega_k) = \mu(\Omega) = 1$. But in fact $\mu(\bigcup_k \omega_k) = \sum \mu(\omega_k) = 0$. The first equality holds because the $\omega_k$'s are disjoint, and the second one because each $\omega_k$ has zero probability of occurring by itself. There is a contradiction, and there exists no measure consistent with the MLN above.[4] The reason the MLN fails to have a measure is that the formulas are not local, in the sense that the truth value of an atom depends on the truth values of infinite others. Locality is in fact the key property for the existence of a consistent measure, and local finiteness ensures it.

**Definition 6.** *A function $f : \Omega \to \mathbb{R}$ is* local *if it depends only on a finite subset $\mathbf{V} \in \mathcal{X}$. A Gibbsian specification $\gamma = (\gamma_{\mathbf{X}})_{\mathbf{X} \in \mathcal{X}}$ is* local *if each $\gamma_{\mathbf{X}}$ is local.*

**Lemma.** *Let $\mathbf{L}$ be a locally finite MLN, and $\gamma^{\mathbf{L}}$ the corresponding specification. Then $\gamma^{\mathbf{L}}$ is local.*

*Proof.* Each Hamiltonian $H_{\mathbf{X}}^{\mathbf{L}}$ is local, since by local finiteness it depends only on a finite number of potentials $\phi_{\mathbf{V}}^{\mathbf{L}}$. It follows that each $\gamma_{\mathbf{X}}^{\mathbf{L}}$ is local, and hence the corresponding specification $\gamma^{\mathbf{L}}$ is also local. $\qquad \square$

We now state the theorem for the existence of a measure admitted by $\gamma^{\mathbf{L}}$.

**Theorem 1.** *Let $\mathbf{L}$ be a locally finite MLN, and $\gamma^{\mathbf{L}} = (\gamma_{\mathbf{X}}^{\mathbf{L}})_{\mathbf{X} \in \mathcal{X}}$ be the corresponding Gibbsian specification. Then there exists a measure $\mu$ over $(\Omega, \mathcal{E})$ admitted by $\gamma^{\mathbf{L}}$, i.e., $|\mathcal{G}(\gamma^{\mathbf{L}})| \geq 1$.*

*Proof.* To show the existence of a measure $\mu$, we need to prove the following two conditions:

1. The net $(\gamma_{\mathbf{X}}^{\mathbf{L}}(\mathbf{X}|\mathbf{S}_{\mathbf{X}}))_{\mathbf{X} \in \mathcal{X}}$ has a cluster point with respect to the weak topology on $(\Omega, \mathcal{E})$.

2. Each cluster point of $(\gamma_{\mathbf{X}}^{\mathbf{L}}(\mathbf{X}|\mathbf{S}_{\mathbf{X}}))_{\mathbf{X} \in \mathcal{X}}$ belongs to $\mathcal{G}(\gamma^{\mathbf{L}})$.

---

[4]See Example 4.16 in Georgii (1988) for a detailed proof.

It is a well known result that, if all the variables $X_i$ have finite domains, then the net in Condition 1 has a cluster point (see Section 4.2 in Georgii (1988)). Thus, since all the variables in the MLN are binary, Condition 1 holds. Further, since $\gamma^L$ is local, every cluster point $\mu$ of the net $(\gamma_X^L(\mathbf{X}|\mathbf{S_X}))_{\mathbf{X}\in\mathcal{X}}$ belongs to $\mathcal{G}(\gamma^L)$ (Comment 4.18 in Georgii (1988)). Therefore, Condition 2 is also satisfied. Hence there exists a measure $\mu$ consistent with the specification $\gamma^L$, as required.                                                                                 □

### 3.3 Uniqueness

This section addresses the question of under what conditions an MLN admits a unique measure. Let us first gain some intuition as to why an MLN might admit more than one measure. The only condition an MLN **L** imposes on a measure is that it should be consistent with the local conditional distributions $\gamma_X^L$. But since these distributions are local, they do not determine the behavior of the measure at infinity. Consider, for example, a semi-infinite two-dimensional lattice, where neighboring sites are more likely to have the same truth value than not. This can be represented by formulas of the form $\forall x, y\ Q(x,y) \Leftrightarrow Q(s(x), y)$ and $\forall x, y\ Q(x,y) \Leftrightarrow Q(x, s(y))$, with a single constant $0$ to define the origin $(0,0)$, and with $s()$ being the successor function. The higher the weight $w$ of these formulas, the more likely neighbors are to have the same value. This MLN has two extreme states: one where $\forall x\ S(x)$, and one where $\forall x\ \neg S(x)$. Let us call these states $\xi$ and $\xi_\neg$, and let $\xi'$ be a local perturbation of $\xi$ (i.e., $\xi'$ differs from $\xi$ on only a finite number of sites). If we draw a contour around the sites where $\xi'$ and $\xi$ differ, then the log odds of $\xi$ and $\xi'$ increase with $wd$, where $d$ is the length of the contour. Thus long contours are improbable, and there is a measure $\mu \to \delta_\xi$ as $w \to \infty$. Since, by the same reasoning, there is a measure $\mu_\neg \to \delta_{\xi_\neg}$ as $w \to \infty$, the MLN admits more than one measure.[5]

Let us now turn to the mathematical conditions for the existence of a unique measure for a given MLN **L**. Clearly, in the limit of all non-unit clause weights going to zero, **L** defines a unique distribution. Thus, by a continuity argument, one would expect the same to be true for small enough weights. This is indeed the case. To make it precise, let us first define the notion of the oscillation of a function. Given a function $f : \mathbf{X} \to \mathbb{R}$, let the oscillation of $f$, $\delta(f)$, be defined as

$$\delta(f) \quad = \quad \max_{\mathbf{x}, \mathbf{x}' \in \mathrm{Dom}(\mathbf{X})} |f(\mathbf{x}) - f(\mathbf{x}')|$$
$$= \quad \max_{\mathbf{x}} |f(\mathbf{x})| - \min_{\mathbf{x}} |f(\mathbf{x})| \qquad (6)$$

---

[5]Notice that this argument fails for a one-dimensional lattice (equivalent to a Markov chain), since in this case an arbitrarily large number of sites can be separated from the rest by a contour of length 2. Non-uniqueness (corresponding to a non-ergodic chain) can then only be obtained by making some weights infinite (corresponding to zero transition probabilities).

The oscillation of a function is thus simply the difference between its extreme values. We can now state a sufficient condition for the existence of a unique measure.

**Theorem 2.** *Let* **L** *be a locally finite MLN with interaction potential* $\Phi^L$ *and Gibbsian specification* $\gamma^L$ *such that*

$$\sup_{X_i \in \mathbf{S}} \sum_{C_j \in \mathbf{C}(X_i)} (|C_j| - 1)|w_j| < 2 \qquad (7)$$

*where* $\mathbf{C}(X_i)$ *is the set of ground clauses in which* $X_i$ *appears,* $|C_j|$ *is the number of ground atoms appearing in clause* $C_j$, *and* $w_j$ *is its weight. Then* $\gamma^L$ *admits a unique Gibbs measure.*

*Proof.* Based on Theorem 8.7 and Proposition 8.8 in Georgii (1988), a sufficient condition for uniqueness is

$$\sup_{X_i \in \mathbf{S}} \sum_{\mathbf{V} \ni X_i} (|\mathbf{V}| - 1)\delta(\Phi_\mathbf{V}^L) < 2 \qquad (8)$$

Rewriting this condition in terms of the ground formulas in which a variable $X_i$ appears (see Definition 3) yields the desired result.                                                        □

Note that, as alluded to before, the above condition does not depend on the weight of the unit clauses. This is because for a unit clause $|C_j| - 1 = 0$. If we define the interaction between two variables as the sum of the weights of all the ground clauses in which they appear together, then the above theorem states that the total sum of the interactions of any variable with its neighbors should be less than 2 for the measure to be unique.

Two other sufficient conditions are worth mentioning briefly. One is that, if the weights of the unit clauses are sufficiently large compared to the weights of the non-unit ones, the measure is unique. Intuitively, the unit terms "drown out" the interactions, rendering the variables approximately independent. The other condition is that, if the MLN is a one-dimensional lattice, it suffices that the total interaction between the variables to the left and right of any arc be finite. This corresponds to the ergodicity condition for a Markov chain.

### 3.4 Non-unique MLNs

At first sight, it might appear that non-uniqueness is an undesirable property, and non-unique MLNs are not an interesting object of study. However, the non-unique case is in fact quite important, because many phenomena of interest are represented by MLNs with non-unique measures (for example, very large social networks with strong word-of-mouth effects). The question of what these measures represent, and how they relate to each other, then becomes important. This is the subject of this section.

The first observation is that the set of all Gibbs measures $\mathcal{G}(\gamma^L)$ is convex. That is, if $\mu, \mu' \in \mathcal{G}(\gamma^L)$ then $\nu \in \mathcal{G}(\gamma^L)$, where $\nu = s\mu + (1-s)\mu'$, $s \in (0, 1)$. This is easily verified by substituting $\nu$ in Equation 4. Hence, the non-uniqueness of a Gibbs measure implies the existence of infinitely many consistent Gibbs measures. Further, many properties of the set $\mathcal{G}(\gamma^{\mathbf{L}})$ depend on the set of extreme Gibbs measures ex $\mathcal{G}(\gamma^{\mathbf{L}})$, where $\mu \in$ ex $\mathcal{G}(\gamma^{\mathbf{L}})$ if $\mu \in \mathcal{G}(\gamma^{\mathbf{L}})$ cannot be written as a linear combination of two distinct measures in $\mathcal{G}(\gamma^{\mathbf{L}})$.

An important notion to understand the properties of extreme Gibbs measures is the notion of a tail event. Consider a subset $\mathbf{S}'$ of $\mathbf{S}$. Let $\sigma(\mathbf{S}')$ denote the $\sigma$-algebra generated by the set of basic events involving only variables in $\mathbf{S}'$. Then we define the tail $\sigma$-algebra $\mathcal{T}$ as

$$\mathcal{T} = \bigcap_{\mathbf{X} \in \mathcal{X}} \sigma(\mathbf{S}_{\mathbf{X}}) \tag{9}$$

Any event belonging to $\mathcal{T}$ is called a tail event. $\mathcal{T}$ is precisely the set of events which do not depend on the value of any finite set of variables, but rather only on the behavior at infinity. For example, in the infinite tosses of a coin, the event that ten consecutive heads come out infinitely many times is a tail event. Similarly, in the lattice example in the previous section, the event that a finite number of variables have the value 1 is a tail event. Events in $\mathcal{T}$ can be thought of as representing macroscopic properties of the system being modeled.

**Definition 7.** *A measure $\mu$ is trivial on a $\sigma$-algebra $\mathcal{E}$ if $\mu(E) = 0$ or 1 for all $E \in \mathcal{E}$.*

The following theorem (adapted from Theorem 7.8 in Georgii (1988)) describes the relationship between the extreme Gibbs measures and the tail $\sigma$-algebra.

**Theorem 3.** *Let $\mathbf{L}$ be a locally finite MLN, and $\gamma^{\mathbf{L}}$ denote the corresponding Gibbsian specification. Then the following results hold:*

1. *A measure $\mu \in$ ex $\mathcal{G}(\gamma^{\mathbf{L}}))$ iff it is trivial on the tail $\sigma$-algebra $\mathcal{T}$.*

2. *Each measure $\mu$ is uniquely determined by its behavior on the tail $\sigma$-algebra, i.e., if $\mu_1 = \mu_2$ on $\mathcal{T}$ then $\mu_1 = \mu_2$.*

It is easy to see that each extreme measure corresponds to some particular value for all the macroscopic properties of the network. In physical systems, extreme measures correspond to phases of the system (e.g., liquid vs. gas, or different directions of magnetization), and non-extreme measures correspond to probability distributions over phases. Uncertainty over phases arises when our knowledge of a system is not sufficient to determine its macroscopic state. Clearly, the study of non-unique MLNs beyond the highly regular ones statistical physicists have focused on promises to be quite interesting. In the next section we take a step in this direction by considering the problem of satisfiability in the context of MLN measures.

## 4 Satisfiability and Entailment

Richardson and Domingos (2006) showed that, in finite domains, first-order logic can be viewed as the limiting case of Markov logic when all weights tend to infinity, in the following sense. If we convert a satisfiable knowledge base $\mathbf{K}$ into an MLN $\mathbf{L_K}$ by assigning the same weight $w \to \infty$ to all clauses, then $\mathbf{L_K}$ defines a uniform distribution over the worlds satisfying $\mathbf{K}$. Further, $\mathbf{K}$ entails a formula $\alpha$ iff $\mathbf{L_K}$ assigns probability 1 to the set of worlds satisfying $\alpha$ (Proposition 4.3). In this section we extend this result to infinite domains.

Consider an MLN $\mathbf{L}$ such that each clause in its CNF form has the same weight $w$. In the limit $w \to \infty$, $\mathbf{L}$ does not correspond to a valid Gibbsian specification, since the Hamiltonians defined in Equation 2 are no longer finite. Revisiting Equation 5 in the limit of all equal infinite clause weights, the limiting conditional distribution is equidistribution over those configurations $\mathbf{X}$ which satisfy the maximum number of clauses given $\mathbf{S_X} = \mathbf{y}$. It turns out we can still talk about the existence of a measure consistent with these conditional distributions, because they constitute a valid specification (though not Gibbsian) under the same conditions as in the finite weight case. We omit the details and proofs for lack of space; they can be found in Singla and Domingos (2007). Existence of a measure follows as in the case of finite weights because of the locality of conditional distributions. We now define the notion of a *satisfying measure*, which is central to the results presented in this section.

**Definition 8.** *Let $\mathbf{L}$ be a locally finite MLN. Given a clause $C_i \in \mathbf{B(L)}$, let $\mathbf{V}_i$ denote the set of Boolean variables appearing in $C_i$. A measure $\mu \in \mathcal{G}(\gamma^{\mathbf{L}})$ is said to be a satisfying measure for $\mathbf{L}$ if, for every ground clause $C_i \in \mathbf{B(L)}$, $\mu$ assigns non-zero probability only to the satisfying assignments of the variables in $C_i$, i.e., $\mu(\mathbf{V}_i = \mathbf{v}_i) > 0$ implies that $\mathbf{V}_i = \mathbf{v}_i$ is a satisfying assignment for $C_i$. $\mathcal{S}(\gamma^{\mathbf{L}})$ denotes the set of all satisfying measures for $\mathbf{L}$.*

Informally, a satisfying measure assigns non-zero probability only to those worlds which are consistent with all the formulas in $\mathbf{L}$. Intuitively, existence of a satisfying measure for $\mathbf{L}$ should be in some way related to the existence of a satisfying assignment for the corresponding knowledge base. Our next theorem formalizes this intuition.

**Theorem 4.** *Let $\mathbf{K}$ be a locally finite knowledge base, and let $\mathbf{L}_\infty$ be the MLN obtained by assigning weight $w \to \infty$ to all the clauses in $\mathbf{K}$. Then there exists a satisfying measure for $\mathbf{L}_\infty$ iff $\mathbf{K}$ is satisfiable. Mathematically,*

$$|\mathcal{S}(\gamma^{\mathbf{L}_\infty})| > 0 \Leftrightarrow Satisfiable(\mathbf{K}) \tag{10}$$

*Proof.* Let us first prove that existence of a satisfying measure implies satisfiability of $\mathbf{K}$. This is equivalent to proving that unsatisfiability of $\mathbf{K}$ implies non-existence of a satisfying measure. Let $\mathbf{K}$ be unsatisfiable. Equivalently, $\mathbf{B}(\mathbf{K})$, the Herbrand base of $\mathbf{K}$, is unsatisfiable. By Herbrand's theorem, there exists a finite set of ground clauses $\mathbf{C} \subseteq \mathbf{B}(\mathbf{K})$ that is unsatisfiable. Let $\mathbf{V}$ denote the set of variables appearing in $\mathbf{C}$. Then every assignment $\mathbf{v}$ to the variables in $\mathbf{V}$ violates some clause in $\mathbf{C}$. Let $\mu$ denote a measure for $\mathbf{L}_\infty$. Since $\mu$ is a probability measure, $\sum_{\mathbf{v} \in \text{Dom}(\mathbf{V})} \mu(\mathbf{V} = \mathbf{v}) = 1$. Further, since $\mathbf{V}$ is finite, there exists some $\mathbf{v} \in \text{Dom}(\mathbf{V})$ such that $\mu(\mathbf{V} = \mathbf{v}) > 0$. Let $C_i \in \mathbf{C}$ be some clause violated by the assignment $\mathbf{v}$ (every assignment violates some clause). Let $\mathbf{V}_i$ denote the set of variables in $C_i$ and $\mathbf{v}_i$ be the restriction of assignment $\mathbf{v}$ to the variables in $\mathbf{V}_i$. Then $\mathbf{v}_i$ is an unsatisfying assignment for $C_i$. Further, $\mu(\mathbf{V}_i = \mathbf{v}_i) \geq \mu(\mathbf{V} = \mathbf{v}) > 0$. Hence $\mu$ cannot be a satisfying measure for $\mathbf{L}_\infty$. Since the above argument holds for any $\mu \in \mathcal{G}(\gamma^{\mathbf{L}_\infty})$, there does not exist a satisfying measure for $\mathbf{L}_\infty$ when $\mathbf{K}$ is unsatisfiable.

Next, we need to prove that satisfiability of $\mathbf{K}$ implies existence of a satisfying measure. We will only give a proof sketch here; the full proof can be found in Singla and Domingos (2007). Let $\mathbf{K}$ be satisfiable. Now, consider a finite subset $\mathbf{X}$ of the variables defined by $\mathbf{L}_\infty$. Given $\mathbf{X}$, let $\Delta_{\mathbf{X}}$ denote the set of those probability distributions over $\mathbf{X}$ which assign non-zero probability only to the configurations which are partial satisfying assignments of $\mathbf{K}$. We will call $\Delta_{\mathbf{X}}$ the set of satisfying distributions over $\mathbf{X}$. $\Delta_{\mathbf{X}}$ is a compact set. Let $\mathbf{Y}$ denote the set of neighbors of the variables in $\mathbf{X}$. We define $F_{\mathbf{X}} : \Delta_{\mathbf{Y}} \to \Delta_{\mathbf{X}}$ to be the function which maps a satisfying distribution over $\mathbf{Y}$ to a satisfying distribution over $\mathbf{X}$ given the conditional distribution $\gamma_{\mathbf{X}}^{\mathbf{L}_\infty}(\mathbf{X}|\mathbf{S}_{\mathbf{X}})$. The mapping results in a satisfying distribution over $\mathbf{X}$ because, in the limit of all equal infinite weights, the conditional distribution over $\mathbf{X}$ is non-zero only for the satisfying assignments of $\mathbf{X}$. Since $\Delta_{\mathbf{Y}}$ is compact, its image under the continuous function $F_{\mathbf{X}}$ is also compact.

Given $\mathbf{X}_i \subset \mathbf{X}_j$ and their neighbors, $\mathbf{Y}_i$ and $\mathbf{Y}_j$ respectively, we show that if $\pi_{\mathbf{X}_j} \in \Delta_{\mathbf{X}_j}$ is in the image of $\Delta_{\mathbf{Y}_j}$ under $F_{\mathbf{X}_j}$, then $\pi_{\mathbf{X}_i} = \sum_{\mathbf{X}_j - \mathbf{X}_i} \pi_{\mathbf{X}_j}$ is in the image of $\Delta_{X_i}$ under $F_{\mathbf{X}_i}$. This process can then be repeated for ever-increasing sets $\mathbf{X}_k \supset \mathbf{X}_i$. This defines a sequence $(\mathbf{T}_i)_{j=i}^{j=\infty}$ of non-empty subsets of satisfying distributions over $\mathbf{X_i}$. Further, it is easy to show that $\forall k \; \mathbf{T}_i^{k+1} \subseteq \mathbf{T}_i^k$. Since each $\mathbf{T}_i^k$ is compact and non-empty, from the theory of compact sets we obtain that the countably infinite intersection $\mathbf{T}_i = \bigcap_{j=i}^{j=\infty} \mathbf{T}_i^j$ is also non-empty.

Let $(X_1, X_2, \ldots, X_k, \ldots)$ be some ordering of the variables defined by $\mathbf{L}_\infty$, and let $\mathbf{X}_k = \{X_1, X_2, \ldots X_k\}$. We now define a satisfying measure $\mu$ as follows. We define $\mu(\mathbf{X_1})$ to be some element of $\mathbf{T}_1$. Given $\mu(\mathbf{X}_k)$, we define $\mu(\mathbf{X}_{k+1})$ to be that element of $\mathbf{T}_{k+1}$ whose marginal is $\mu(\mathbf{X}_k)$ (such an element always exists, by construction). For an arbitrary set of variables $\mathbf{X}$, let $k$ be the smallest index such that $\mathbf{X} \subseteq \mathbf{X}_k$, and define $\mu(\mathbf{X}) = \sum_{\mathbf{X}_k \setminus \mathbf{X}} \mu(\mathbf{X}_k)$. We show that $\mu$ defined in such a way satisfies the properties of a probability measure (see Section 2.2). Finally, $\mu$ is a satisfying measure because $\forall k \, \mu(\mathbf{X}_k) \in \mathbf{T}_k$ and each $\mathbf{T}_k$ is a set of satisfying distributions over $\mathbf{X}_k$. □

**Corollary.** *Let $\mathbf{K}$ be a locally finite knowledge base. Let $\alpha$ be a first-order formula, and $\mathbf{L}_\infty^\alpha$ be the MLN obtained by assigning weight $w \to \infty$ to all clauses in $\mathbf{K} \cup \{\neg \alpha\}$. Then $\mathbf{K}$ entails $\alpha$ iff $\mathbf{L}_\infty^\alpha$ has no satisfying measure. Mathematically,*

$$\mathbf{K} \models \alpha \;\Leftrightarrow\; |\mathcal{S}(\gamma^{\mathbf{L}_\infty^\alpha})| = 0 \qquad (11)$$

Thus, for locally finite knowledge bases with Herbrand interpretations, first-order logic can be viewed as the limiting case of Markov logic when all weights tend to infinity. Whether these conditions can be relaxed is a question for future work.

## 5 Related Work

A number of relational representations capable of handling infinite domains have been proposed in recent years. Generally, they rely on strong restrictions to make this possible. To our knowledge, Markov logic is the most flexible language for modeling infinite relational domains to date. In this section we briefly review the main approaches.

Stochastic logic programs (Muggleton, 1996) are generalizations of probabilistic context-free grammars. PCFGs allow for infinite derivations but as a result do not always represent valid distributions (Booth & Thompson, 1973). In SLPs these issues are avoided by explicitly assigning zero probability to infinite derivations. Similar remarks apply to related languages like independent choice logic (Poole, 1997) and PRISM (Sato & Kameya, 1997).

Many approaches combine logic programming and Bayesian networks. The most advanced one is arguably Bayesian logic programs (Kersting & De Raedt, 2001). Kersting and De Raedt show that, if all nodes have a finite number of ancestors, a BLP represents a unique distribution. This is a stronger restriction than finite neighborhoods. Richardson and Domingos (2006) showed how BLPs can be converted into Markov logic without loss of representational efficiency.

Jaeger (1998) shows that probabilistic queries are decidable for a very restricted language where a ground atom cannot depend on other groundings of the same predicate. Jaeger shows that if this restriction is removed queries become undecidable.

Recursive probability models are a combination of Bayesian networks and description logics (Pfeffer & Koller, 2000). Like Markov logic, RPMs require finite neighborhoods, and in fact existence for RPMs can be proved succinctly by converting them to Markov logic and applying Theorem 1. Pfeffer and Koller show that RPMs do not always represent unique distributions, but do not study conditions for uniqueness. Description logics are a restricted subset of first-order logic, and thus MLNs are considerably more flexible than RPMs.

Contingent Bayesian networks (Milch et al., 2005) allow infinite ancestors, but require that, for each variable with infinite ancestors, there exist a set of mutually exclusive and exhaustive contexts (assignments to finite sets of variables) such that in every context only a finite number of ancestors affect the probability of the variable. This is a strong restriction, excluding even simple infinite models like backward Markov chains (Pfeffer & Koller, 2000).

Multi-entity Bayesian networks are another relational extension of Bayesian networks (Laskey & Costa, 2005). Laskey and Costa claim that MEBNs allow infinite parents and arbitrary first-order formulas, but the definition of MEBN explicitly requires that, for each atom $X$ and increasing sequence of substates $S_1 \subset S_2 \subset \ldots$, there exist a finite $N$ such that $P(X|S_k) = P(X|S_N)$ for $k > N$. This assumption necessarily excludes many dependencies expressible in first-order logic (e.g., $\forall x \; \exists! y \; \texttt{Loves(y, x)}$). Further, unlike in Markov logic, first-order formulas in MEBNs must be hard (and consistent). Laskey and Costa do not specify a language for specifying conditional distributions; they simply assume that a terminating algorithm for computing them exists. Thus the question of what infinite distributions can be specified by MEBNs remains open.

## 6   Conclusion

In this paper, we extended the semantics of Markov logic to infinite domains using the theory of Gibbs measures. We gave sufficient conditions for the existence and uniqueness of a measure consistent with the local potentials defined by an MLN. We also described the structure of the set of consistent measures when it is not a singleton, and showed how the problem of satisfiability can be cast in terms of MLN measures. Directions for future work include designing lifted inference and learning algorithms for infinite MLNs, deriving alternative conditions for existence and uniqueness, analyzing the structure of consistent measure sets in more detail, extending the theory to non-Herbrand interpretations and recursive random fields (Lowd & Domingos, 2007), and studying interesting special cases of infinite MLNs.

## Acknowledgements

We are grateful to Michael Jordan for helpful discussions. This research was partly funded by DARPA contract NBCH-D030010/02-000225, DARPA grant FA8750-05-2-0283, NSF grant IIS-0534881, and ONR grant N-00014-05-1-0313. The views and conclusions contained in this document are those of the authors and should not be interpreted as necessarily representing the official policies, either expressed or implied, of DARPA, NSF, ONR, or the United States Government.